# The pursuit of beauty: Converting image labels to meaningful vectors

Savvas Karatsiolis, Andreas Kamilaris
Research Centre on Interactive Media, Smart Systems and Emerging Technologies
(RISE), Nicosia, Cyprus
s.karatsiolis@rise.org.cy, a.kamilaris@rise.org.cy

[ABSTRACT]

A challenge of the computer vision community is to understand the semantics of an image, in order to allow image reconstruction based on existing high-level features or to better analyze (semi-)labelled datasets. Towards addressing this challenge, this paper introduces a method, called Occlusion-based Latent Representations (OLR), for converting image labels to meaningful representations that capture a significant amount of data semantics. Besides being informational rich, these representations compose a disentangled low-dimensional latent space where each image label is encoded into a separate vector. We evaluate the quality of these representations in a series of experiments whose results suggest that the proposed model can capture data concepts and discover data interrelations.

## 1. Introduction

Understanding the semantics or the high-level features of an image is an important step in better analyzing labelled or semi-labelled datasets, as well as to reconstructing images based on semantic information available in the datasets. This could lead to the next generation of image filtering, with applications in online social media (e.g. changing human faces), to landscapes and architecture etc. Imagine a landscape of a lake with a mountain in the background, which could be enhanced with a small island in the middle of the lake. Or a woman dressed in particular clothes and then the model adding automatically the shoes that fit better to her.

Deep learning (DL) advancements during the last years offer powerful frameworks for mapping dataset instances to binary labels and thus allow the building of powerful classifiers for a number of seemingly difficult tasks [1,2,3,4]. Classification models are usually simpler and more successful in their task than generative models. Transforming the instances of a dataset to meaningful representations is harder than transforming them to binary vectors because it requires the preservation of data semantics. Compressing the dataset instances into binary numbers forces the condensation of data interrelations to a degree that they become undetectable and not re-constructible anymore. For example, the value of a binary label can be calculated by a classifier by combining the features detected during forward propagation of the data through the model's layers. At the level where the label is calculated (i.e. the model's output) every feature and data characteristic has already been processed in some high-level data abstraction. More importantly, the high-level concepts do not contain qualitative information or any other statistical info that describes the degree based on which some instance complies with a specific label. On the contrary, a label described by a distribution instead of a binary label mitigates the problem of blurred-out statistics and context unawareness.

The contribution of this paper involves the investigation of suitable DL models which allow the calculation of meaningful vectors from the labels of a dataset with information provided by a classifier trained with these labels. For this purpose, a Siamese neural network is employed, where he information of the classifier combined with input image occlusion enables the Siamese model to extract discriminating features and calculate meaningful label distributions. We call our method Occlusion-based Latent Representations (OLR). This work's main contribution is a simple but effective methodology for learning appropriate label distributions that contain enough semantic information and can be exploited in various ways as demonstrated in a series of experiments. OLR builds latent representations in a supervised manner (using labels) that have a major advantage: latent subspaces disentanglement. Each latent representation links directly to one problem label and automatically constitutes that label's exclusive factor of variation.

The rest of the paper is organized as follows: Section 2 describes related work, Section 3 describes the methodology followed while Section 4 presents various experiments assessing the quality and effectiveness of the approach. Then, Section 5 discusses the results and Section 6 concludes the paper.

## 2. Related Work

Different studies applied a variety of strategies for label enhancement and/or efficient separation of the effect that each label has on the data.

### 2.1 Label distribution learning and Label enhancement

Hinton et al [5] suggested raising the temperature coefficient of the SoftMax units at the output of the classifier in order to increase the entropy of the label distribution. In this way, label distribution becomes less stringent and reveals otherwise unobservable instance properties. Of course, this strategy can be applied only after the model's training and it allows instance representation with relaxed class probabilities, i.e. a vector containing the probability of each class. Label distribution learning (LDL) [6] aimed to a similar outcome, i.e. a vector representing the degree to which each label describes an instance. LDL mapped each instance to a label distribution space but required the availability of the actual label distribution before-hand, something which is highly impracticable in real-world applications. Label enhanced multi-label learning (LEMLL) [7] suggested a framework incorporating regression of the numerical labels and label enhancement that did not require the availability of the label distribution. LEMLL jointly learned the numerical labels and the predictive model taking advantage of the topological structure in the feature space (label enhancement).

### 2.2. Attribute-editing models

Attribute editing models are also relevant, in the sense that they target specific attributes of the instances. Some recent attribute-editing models manipulated face attributes and generated images with a set of desired (or undesired) characteristics while preserving at the same time almost all other image details. Given an image and the desired characteristics (labels), an image was generated that satisfied the given characteristics resembling the initial image in every other detail. Fader networks [8] enforced an adversarial process that makes the latent space of the labels invariant to the attributes. Generally, the attribute-independent latent representation is very restrictive, leading to blurriness and distortion [8]. Wei Shen and Rujie Liu [9] proposed a model that learns the difference between images before and after the manipulation, i.e. a residual image holding the difference of the pixel values due to attribute-editing.

An interesting approach applying an encoder-decoder architecture is MulGan [10], which compressed the original image to a latent representation that had predefined placeholders for the different problem classes. The model was trained using different image pairs, editing the individual placeholders according to the corresponding binary labels of each image, preserving only the ones that are set in the image label vector and making zero every other. Edited representations passed through the decoder to reconstruct the original image. Besides the two edited representations, MulGan created two more representations by exchanging the editable placeholders between the two representations. The representations with the attributes exchanged passed through a label classifier and a real/fake discriminator. The latter used an adversarial loss aiming to produce more realistic images. AttGan [11] also used an encoder-decoder architecture but additionally applied a conditional decoding of the latent representation based on the desired attributes (i.e. class labels). AttGan also applied a reconstruction, using both a classification and an adversarial loss. The reconstruction preserved the attribute-excluding details, classification loss guaranteed correct attribute manipulation while adversarial loss aimed to achieve realistic image generation. Authors of AttGan also suggested that symmetric skip connections between the encoder and the decoder, similar to the U-Net architecture [12], improved their model's performance. STGAN [13] made some significant modifications to the AttGan architecture for further improving the results obtained. The authors of STGAN, after conducting several experiments, suggested that skip connections can improve the reconstruction of the original image but at the same time may harm attribute-editing. Their effect can be driven to a win-win compromise by the use of selective transfer units that control the information flow from the encoder to the decoder. They also suggested using a difference attribute vector instead of the whole actual target attribute vector. By only using the desired differences vector (having a *-1* for removing an attribute and a *+1* for adding an attribute), reconstruction quality was improved.

### 2.3. Disentangled representations

According to Bengio et al [25], a change in one dimension of a disentangled representation causes a change in one variation factor while being relatively invariant to changes in other factors. Disentangled representations have been studied both in the context of semi-supervised learning [26, 27] and unsupervised learning [28, 29, 30]. Semi-supervised approaches require knowledge about the underlying factors of the data which is a significant limitation. β-VAE [28] is a disentangling approach based on the Variational Autoencoder (VAE) [23] and achieves latent space disentanglement by applying a slightly different VAE objective function with a larger weight on the Kullback–Leibler (KL) divergence between the posterior and the prior. While the β-VAE is appealing mainly because it relies on the elegant framework of the VAE, it offers disentanglement to the cost of generated image quality. Kim and Mnih [30] proposed encouraging the VAE's representations distribution to be factorial. which improves upon β-VAE. InfoGAN [29] is a popular alternative that enhances the mutual information between the latent codes and the generated images.

## 3. Methodology

Our approach for turning the problem labels to distributions involves the use of information from a model trained on the classification task. Such a classifier compresses the information of an image down to labels and outputs probabilities of label occurrence for an input image. We further use a Siamese network [14, 15, 16, 17] which receives two images and the product of their label probabilities to adapt its output accordingly, as illustrated in Figure 2. The output of the Siamese network $E \in \mathcal{R}^{L \times k}$ comprises of $L$ vectors of size $k$, with $L$ being the number of problem labels and $e^l \in \mathcal{R}^k$ being the row vector component of $E$ corresponding to label $l$. Effectively, the Siamese output is a matrix holding much less information than the original input $x \in \mathcal{R}^{h \times w \times 3}$, where $h, w$ are the height and width of the 3 channels of the image respectively. We generally assume that $L \times k \ll h \times w \times 3$. The output consists of $L$ distributions in vector form, one for each problem label. Since these vectors constitute compressed representations of the input, we will refer to them as *image embeddings* from this point on.

For the Siamese model to learn the embeddings properly, we sample pairs of images from the dataset calculating their label probabilities using a classifier previously trained on recognizing the labels. The probability outcomes are multiplied in an element-wise fashion in order to obtain a value for the overall probability of each label being evident in the image pair. Each training example comprises of a triplet of two images and the joint probability vector of the problem labels (the product of the classifier's probabilities). The Siamese network receives the two images of each triplet and calculates two embeddings, one for each image. Then, it calculates the dot products between the vector components $e^l$ of the two embeddings. Assuming the two embeddings matrices $E_1, E_2 \in \mathcal{R}^{L \times k}$, the dot product is calculated between the rows of the two matrices resulting in a vector $\vec{v} \in \mathcal{R}^L$. The loss function of the Siamese model is equal to the Mean Squared Error (MSE) between $\vec{v}$ and the joint probability vector in the triplet as defined above. In other words, the dot product between the label embeddings of the two images should be equal to the joint probability vectors as calculated by the classifier outcomes. This means that images that share a common label should have class embeddings with a high dot product.

Regarding the proposed approach, there are two main issues to address. The first has to do with the Siamese model architecture and the way it is designed to have an output in the form of matrix $E$. The second issue concerns the calculation of appropriate joint-probability vectors. Regarding the Siamese architecture, after several convolutional and pooling layers, we apply a special layer that comprises of several feature maps that form label-specific groups. The number of groups is equal to the number of problem labels, so that each label is represented by a certain number of feature maps. The number of feature maps representing a label is equal to the dimensionality of the label vectors $k$ and the size of the special layer is $f \times f \times (L \times k)$, with $f$ being the width/height of the feature maps. At the output of this layer, an average pooling layer is applied which calculates the average of each feature map. Consequently, the output of the average pooling layer is $L \times k \times 1$ and, through a reshaping operation, the output can be transformed to the embedding's shape $L \times k$. The last layers of the Siamese network are displayed in Figure1.

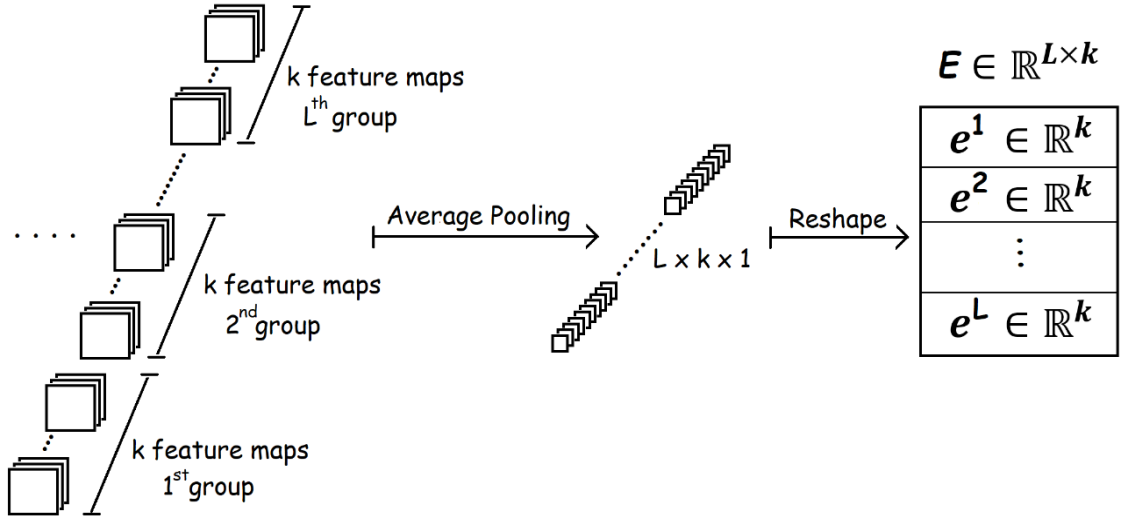

Figure 1. Final layers of the Siamese model. After several convolutional and pooling layers, the label-specific layer consists of $L$ groups of $k$ feature maps. The next layer is an average pooling layer followed by a reshape operation which formats the output matrix to a shape of $L \times k$, so that there is a vector (embedding) of size $k$ for each of the $L$ classes. According to the architecture, each problem label has its own individual feature maps which represent its statistics, providing its $k$ components.

The concern for calculating appropriate joint-probability vectors has to do with the classifier's tendency to output a high probability (close to *1*) for the correct class and a low probability (close to *0*) for incorrect classes. This results in calculating joint probabilities that do not empower the Siamese model to learn the data interrelations. The Siamese model becomes inefficient when its training relies on over-confident vectors or vectors of binary nature. Additionally, when joint probabilities lie close to the extreme probability values (*0* or *1*) the Siamese model is more prone to overfitting and thus may not properly consider feature correlations and interactions. Two ways for raising the entropy of the classifier's output were considered: a) The first uses model distillation [5] by raising the temperature parameter of the SoftMax function at the output of the classifier which relaxes the label distribution and communicates more information about the input; b) The second approach is based on applying random partial occlusion to the input in order to make the classifier less confident about its predictions.

Experiments showed that occlusion works better in the sense that it prevents the Siamese model from overfitting and encourages the discovery of feature correlations and the calculation of more expressive distributions. The degree of the occlusion on the images of each triplet (the percentage of the occluded image surface) can be determined experimentally for the problem at hand. We discovered that randomly selected rectangles of width and height ranging from $33\% - 66\%$ of the image dimensions have a positive effect on the training of the Siamese model. Figure 2 shows the training process of the Siamese model.

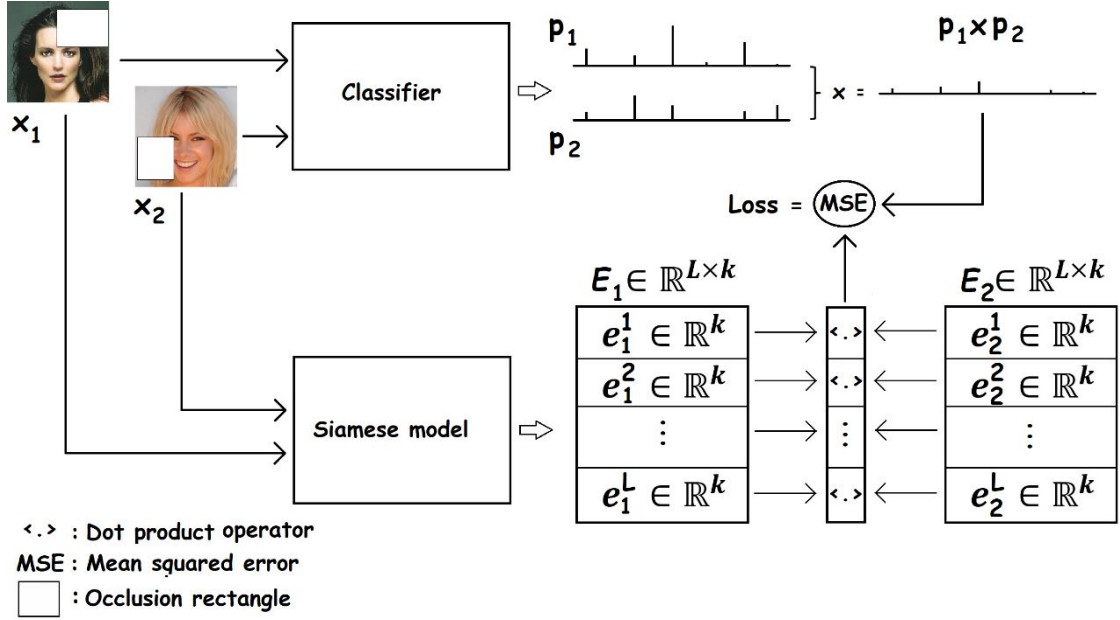

Figure 2. Training process of the Siamese model. Two images are randomly chosen from the dataset and occlusion rectangles are applied at random positions on them. These rectangles are of different shapes and have a height and width randomly chosen from a range of values that are between $\frac{1}{3}$ and $\frac{2}{3}$ of the image height and width (the occlusion rectangles shown in the figure are smaller in size for cosmetic reasons). Next the two occluded images are classified, and the resulting label probabilities are multiplied to form a joint label probability. The two occluded images and the joint probability vector form a triplet that is used to train the Siamese model. Both images go through the model and produce two image embeddings $E_1, E_2$. A dot product operation applies to the vector components of the embeddings resulting to a vector of $L$ elements. The MSE between the joint probability of the triplet and the dot product vector constitutes the loss of the training procedure.

## 4. Experimental results

We evaluate the proposed method on the CelebA dataset [18]. The dataset contains more than 200,000 images of faces, each annotated with 40 binary labels (either an attribute exists or not). Images are cropped and resized to $178 \times 178 \times 3$ pixels. In several cases, cropping removes part of the person's neck thus 2 labels requiring view on the specific (low) image region are not considered: "*wearing necklace*" and "*wearing necktie*". This reduces the number of problem labels to *38*. Randomly selected 190,000 images are used in the training set (95%) while the 10,000 remaining images are kept for the test set (5%). No pre-processing has been applied to the images. After training the model and calculating the embeddings for each image in the dataset their quality is evaluated through various experiments discussed in the following sub-sections.

### 4.1 Experiment 1: Using the embeddings to train a linear classifier

A linear classifier was trained based on the CelebA dataset using the calculated embeddings, to assess their quality. The performance of the convolutional classifier that the Siamese model relies on for its training was used as a baseline. This comparison can provide some useful insights on whether the calculated embeddings are indeed capturing the data relations.

The linear classifier for this experiment has a single layer comprising of *38* neurons representing the classes of the dataset. Each of these neurons uses the sigmoid activation. The embeddings calculated by the Siamese model $E \in \mathbb{R}^{N \times 38 \times 32}$ (*N* being the size of the training set) are used as input to this linear model. The linear classifier has a classification success rate of 91.6% on the test set while the convolutional classifier has a success rate of 94.2%. This slight performance decrease is the cost of obtaining embeddings that capture data inter-relations, as will be shown briefly.

The patterns classified incorrectly by the linear model (trained on the embeddings) but, at the same time, classified correctly by the convolutional classifier were further analyzed. These cases belong to 2.6% of the test set that reflects the success rate difference of the classifiers in comparison. It turns out that the CelebA dataset contains several wrong or ambiguous labels that the Siamese embeddings did not agree with. Some examples of questionable cases are shown in Figure 3. The Siamese model seems to be reluctant to associate vague labels with false evidence (features). On the contrary, the convolutional classifier tends to adopt the ambiguous labels acting obediently in an eager-to-satisfy fashion.

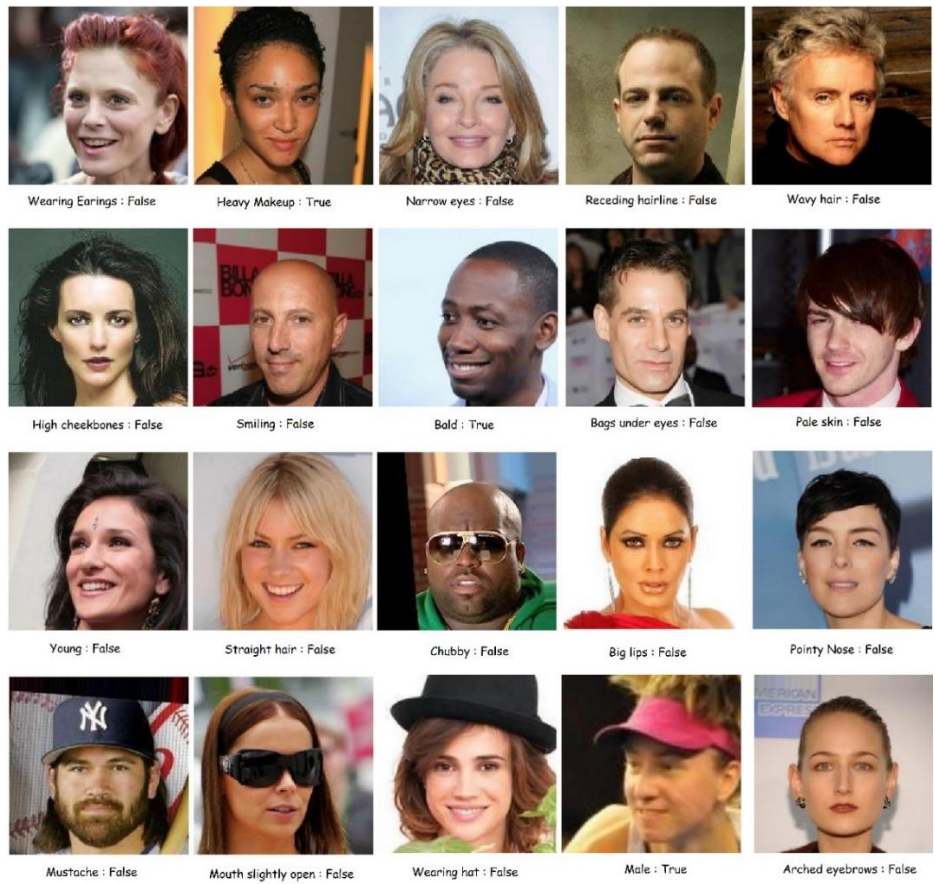

Figure 3. Examples of original CelebA images that are accompanied with vague labels, shown under the images. OLR does not adopt this labeling. Generally, the Siamese embeddings are more resilient to such cases than a convolutional classifier in the sense that they adopt a label only if they discover strong feature correlations with other images having the corresponding label.

## 4.2 Experiment 2: Correlations between the embeddings' distributions

In this experiment, the correlations between the embeddings were examined to investigate empirically whether the depicted label distributions are rational. Figure 4 shows the Pearson correlation coefficients between the distributions' norm value. When a label is detected, the corresponding embedding's norm-value tends to increase reflecting the presence of such a characteristic, otherwise the norm-value is very small. The high values of the vector reveal an attempt to describe the evident label through the calculated distribution. Some interesting and well anticipated correlations are revealed, such as for example the positive effect that big lips (*0.3*) and wearing lipstick (*0.7*) may have on considering a person being attractive.

Other interesting correlations are between baldness and attractiveness (*-0.2*), double chin and gray hair (*+0.5*), being young and bald (*-0.3*), high cheek bones and attractiveness (*0.3*), being male and having a big nose (*0.6*), being male and having a heavy makeup (*-0.8*) and the tendency to consider a smiling person attractive (*0.2*). Small steps towards the pursuit of beauty are being made here.

## 4.3 Experiment 3: Principal component analysis of the embeddings' distribution

Principal component analysis (PCA) was applied to the embeddings focusing on the label "*Mouth slightly open*", in order to further analyze the results and evaluate the characteristics of the distribution as obtained from OLR. This specific label was selected because almost half of the images in the dataset contain it, hence there is much information available for analysis. Moreover, this label can be effortlessly detected in an image and its detection does not rely on subjective judgment, such as for example the label regarding "*attractiveness*". The PCA applied on the "*Mouth slightly open*" embeddings of all images in the dataset revealed that the first component (eigenvector) explains 67.5% of the data variance while the second component explains another 4.1% of the data variance. Given the large quantity of variance explained by the first component, only this component was selected in this experiment. The projections of all the embeddings on this single component were sorted in an increasing order of magnitude, viewing the images corresponding to several locations in the ordered list, which has a size equal to the whole dataset (~204,000). We started from images having smaller projection on the first principal component moving towards images that have a larger projection and thus comply with the selected label "*Mouth slightly open*". Figure 5 shows images from the first principal component ranking.

A higher value of the principal component projection signifies more confidence in the label "*open mouth*" being evident. While this is true for the images of the second row, where the faces possess the specific attribute, the increase of the projection value is not reflected on the images of the first row (where the faces do not comply with the specific label): the increase of the principal component projection does not translate to a less shut mouth. This inconsistency occurs because the dot product function imprints the degree of similarity between the images that share a specific attribute and so depicts a small embedding norm when the images do not share the attribute. While image similarity in terms of common labels is based on probabilistic targets of some value, image dissimilarity relies on a dot product target value which is zero or very close to zero.

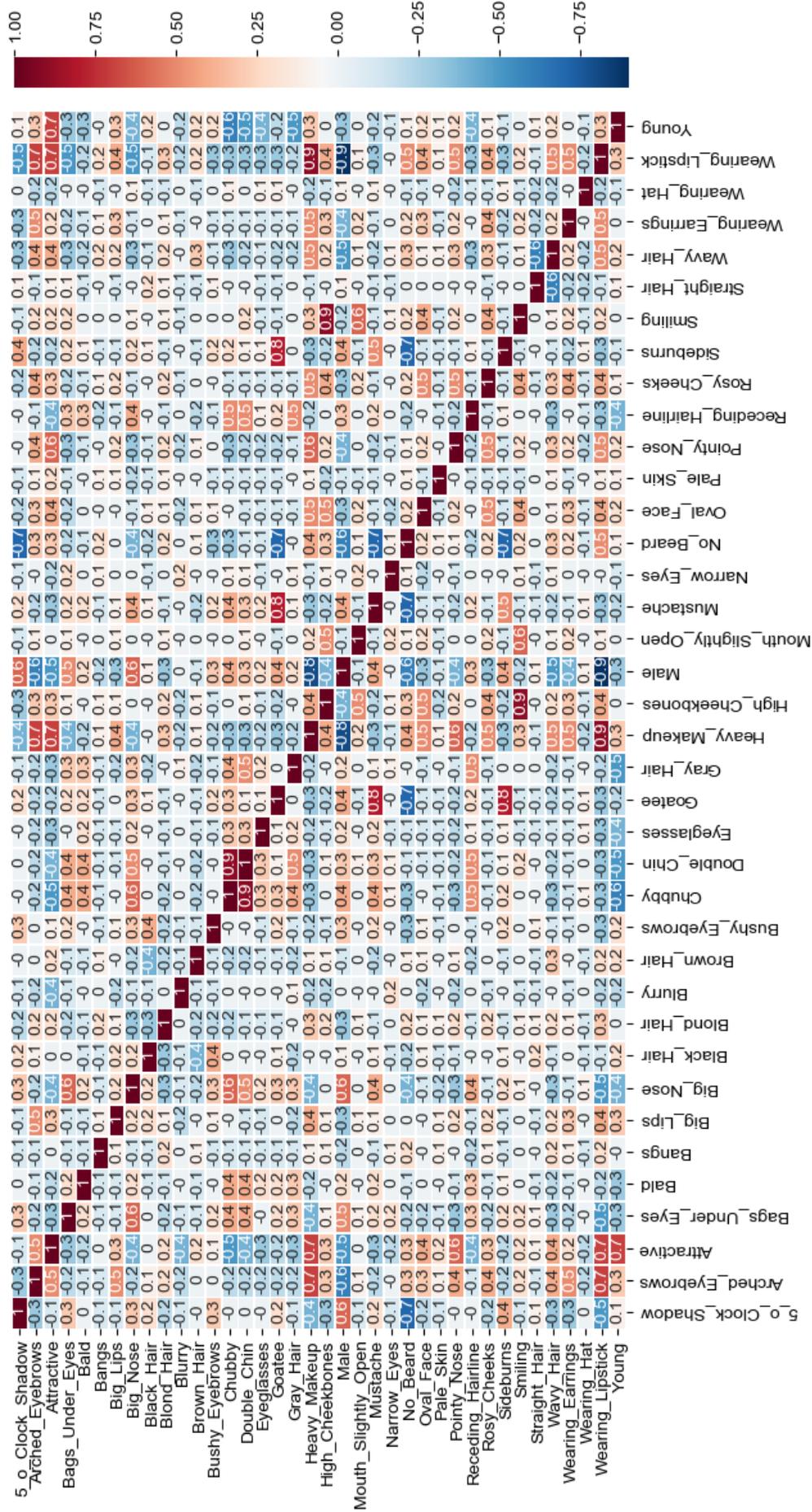

Figure 4. The Pearson correlation table between the labels' vectors norm value. The norm value of a specific label vector tends to increase when a positively correlated label is evident in the same pattern. Likewise, it tends to decrease at the presence of a negatively correlated label.

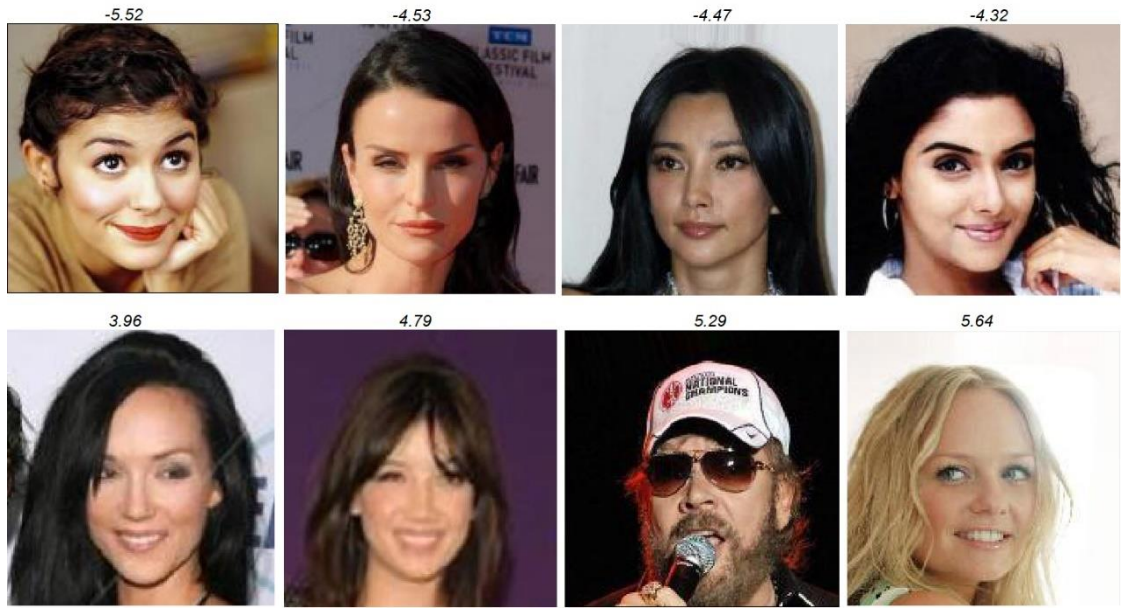

Figure 5. Images corresponding to different projection values of the "*Mouth slightly open*" embeddings on the first principal component of the specific label embeddings set (displayed in an increasing order). In the top row, the images correspond to ranking locations which are 2*0,000* positions apart (ranking positions *0-80,000*). From that point on, images satisfy the "*Mouth slightly open*" attribute (almost 50% of the images in the dataset have the specific label). The second row shows images corresponding to the ranking positions *100,000-180,000*. The actual projection value is shown on top of each image.

**4.4 Experiment 4: Using the embeddings for reconstructing the images**

In this experiment, an embedding is calculated for each of the problem labels. If a label is evident in an image, its corresponding vector output imprints the phenotype of the specific label in the image. Each dataset image has an average of eight non-zero labels, which means that the average embeddings' size effectively describing an image is $8 \times 38 = 256$ out of the total $32 \times 38 = 1216$ numbers of the model's output. The validity of this analysis is based on the fact that any label not evident in an image is described with a zero (or near zero) vector, so only the active labels get a non-zero vector value. Given the input image sizes ($178 \times 178 \times 3 = 95052$), the model compresses the input by more than 370 times, representing the images with only $\bar{k} \times 32$ numbers, where $\bar{k}$ is the average number of evident labels in the images (non-zero labels). Due to the huge compression rate, reconstructing the image in a way that the imprinted face is recognized as being the same face shown in the input image is a challenging task.

The MSE loss for the image reconstruction process has some interesting properties but also tends to create blurry images and annoying artifacts [19, 20]. The very large compression factor applied in the embeddings amplifies these disadvantages. The MSE or any other norm-based distance error does not account for the structure and the characteristics of an image, such as the statistics among pixel values. On the contrary, such losses produce reconstructions which, in the general case, only approximate the raw pixel values in the training images. A better reconstruction could be obtained by using a loss function that accounts for pixel statistics reflecting the structure of the images like the structural similarity loss function (SSIM) [21]. The SSIM loss function considers three basic image components: luminance, contrast and structure. SSIM is a perceptual-based loss function which considers some factors that are closer to what humans perceive when they look at an image. It seems unlikely that humans perceive an image's content by making pixel-level calculations in a way similar to what norm-based losses do. In practice, the SSIM loss function measures the similarity between two images based on factors that encode the perceived change in structural information. The SSIM of two images $x, y$ is calculated on various windows of size $N$ as follows:

$$SSIM(x,y) = \frac{(2\mu_x\mu_y + c_1)(2\sigma_{xy} + c_2)}{(\mu_x^2 + \mu_y^2 + c_1)(\sigma_x^2 + \sigma_y^2 + c_2)}$$

$$\mu_x = \frac{1}{N}\sum_N x_i \;,\; \mu_y = \frac{1}{N}\sum_N y_i$$

$$\sigma_x^2 = \frac{1}{N-1}\sum_N (x_i - \mu_x)^2 \;,\; \sigma_y^2 = \frac{1}{N-1}\sum_N (y_i - \mu_y)^2$$

$$\sigma_{xy} = \frac{2\sigma_x \sigma_y + c_2}{\sigma_x^2 + \sigma_y^2 + c_2}$$

where $c_1$ and $c_2$ are two variables used to stabilize the division.

SSIM acts on the luma (brightness) of the images and does not consider chrominance. For that reason, SSIM is applied separately on each of the three-color channels of the image. In order to achieve chromatic reconstruction, the MSE loss was also used in conjunction with the SSIM, in a way that allows relative freedom to each loss function's application. This degree of freedom is accomplished by applying the two losses on different layers of the reconstruction model, allowing both SSIM and MSE to operate on different value scales. More specifically, the SSIM loss is applied first to the output of the second-last layer $q^{-1}$ of the model, which has a size of $176 \times 176 \times 3$ as shown in Table 1. A pixel was removed from each side (top/bottom height and left/right width) of the dataset images to match the size of the model's output. Next, a normalizing layer (last layer) $q$ puts the pixel values back to the range $y \in [0,1]$ by applying the following operation on the output of the previous layer $x$:

$$y_i = \frac{x_i - \min(x)}{\max(x) - \min(x)}$$

Then, the MSE loss is applied at the last layer after the SSIM loss is scaled by a factor $a$. The total loss function between the reconstructed image $y$ and the original image $x$ that corresponds to an input embedding $E$ is:

$$L_e = a\, SSIM(x, q^{-1}) + (1-a) MSE(x, q)$$

Table 1. The architecture of the decoder model. Parameters are in the form (*k, k, c, s, p*) representing the kernel size *(k)*, channels number *(c)*, strides *(s)* and padding *(p)*.

| Layer | Output Size | Parameters |
|---|---|---|
| Flatten | (1216) | |
| Dense | (18432) | |
| Reshape | (6,6,512) | |
| Conv2DTranspose | (12,12,128) | (3,3,128,2,'same') |
| Conv2D | (12,12,64) | (3,3,64,1,'same') |
| Conv2DTranspose | (24,24,64) | (3,3,64,2,'same') |
| Conv2D | (22,22,64) | (3,3,64,1,'valid') |
| Conv2DTranspose | (44,44,64) | (3,3,64,2,'same') |
| Conv2D | (44,44,64) | (3,3,64,1,'same') |
| Conv2DTranspose | (88,88,64) | (3,3,64,2,'same') |
| Conv2D | (88,88,64) | (3,3,64,1,'same') |
| Conv2DTranspose | (176,176,64) | (3,3,64,2,'same') |
| Conv2D | (176,176,32) | (3,3,32,1,'same') |
| Conv2D_Out | (176,176,3) | (3,3,3,1,'same') |
| SSIM(Image, Conv2D_Out) | 1 | |
| Rescale(Conv2D_Out) | (176,176,3) | |
| MSE(Image, Rescale) | 1 | |

The reconstruction model is shown in Table 1 and it is trained with the RMSProp optimizer and a learning rate of $1 \times e^{-4}$. Some reconstructions based on the test set embeddings are shown in Figure 6 next to the original images that produced the embeddings. The reconstructions suggest that the embeddings hold significant information from the original images, despite the huge compression. More specifically, the reconstructions generally tend to preserve the general facial structure, individual characteristics, pose and facial expressions. This behavior is interesting for the following reasons:

1. The reconstruction model and the embeddings-extracting model are separately trained with different objective functions and there is no co-adaptation of their individual tasks. However, these models can be joined together to form an implied under-determined auto encoder that significantly compresses the original image to a small internal representation (embedding) and then decode it to reconstruct the original image.
2. All images illustrated in Figure 6 belong to the test set which means that the models (embeddings' extraction model and decoder) have never seen them before. These images were not used during the training of neither model.

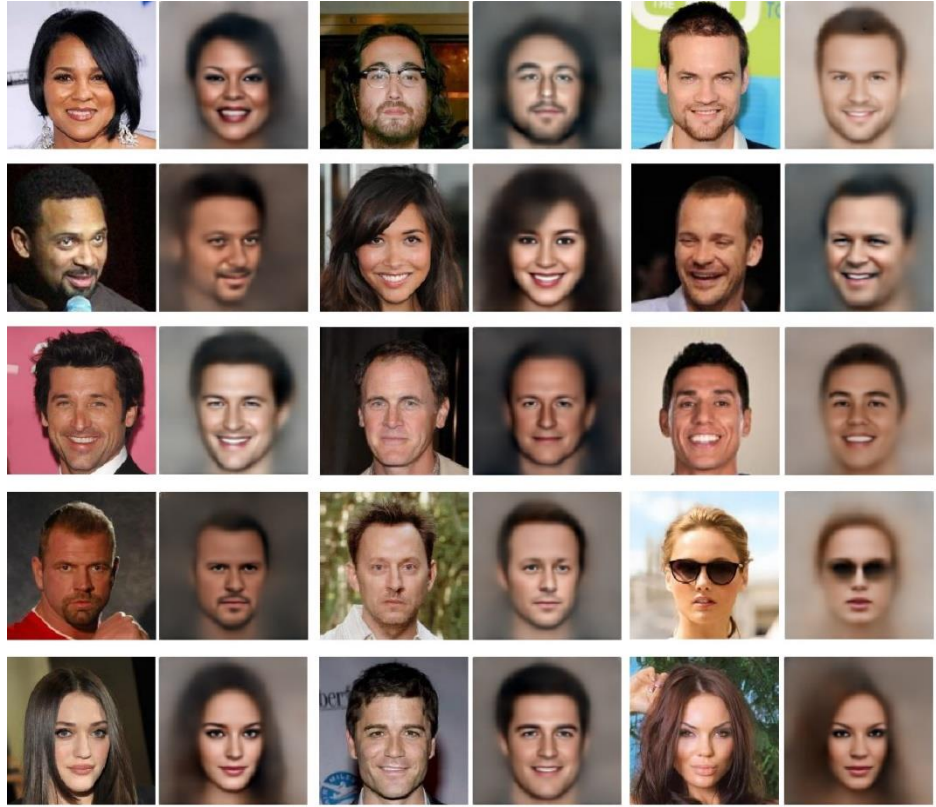

Figure 6. Several reconstructions of the decoder model in comparison to the original images that produced the embeddings used as input to the decoder model. Each pair of images consists of the original image on the left and the reconstruction on the right (3 pairs per row). Most reconstructions tend to preserve the facial structure and characteristics, pose and facial expression information. The original images belong to the test set and have not been used during training of the Siamese model or the decoder model.

It must be noted that the embeddings-extracting model calculates vectors containing significantly less information than the original image (~370 times less information). The model maintains an important degree of similarity between the reconstructions and the original images despite the high compression ratio.

**4.5 Experiment 5: Using the image embeddings for image editing**

As long as the calculated embeddings are converted to a distribution (or distributions of the various problem labels), we can extract and apply inferred statistical properties on the data. More specifically, it is assumed that the *32-dimensional* (32-d) vectors representing a specific class (out of the *38* dataset classes) are points on a normal probability distribution. Then, each of these *32-d* vectors corresponding to a specific class is selected to calculate the distribution function of the specific problem label based only on the data that corresponds to that class label. For example, the distribution of the "*wearing eyeglasses*" class is calculated from a group containing the dataset patterns satisfying the specific label. Let $E_i$ be the embedding of an image having a size of $38 \times 32$ and $e_i^l$ be the *32-d* vector component that corresponds to a single class $l$ out of the *38* classes described by the embedding. The mean of the distribution formed by the class $l$ vector components $e^l$ of all $N$ images in the dataset that satisfy the specific label $\left(y_i^l = 1\right)$ is calculated by

$$\mu_l = \frac{1}{\sum_{i,y_i^l=1}^N 1} \sum_{i,y_i^l=1}^N e_i^l.$$

The covariance matrix of the normal distribution $\Sigma \in R^{k \times k}$, where $k$ is the vector dimensionality (*32*), is calculated in a matrix form with:

$$\Sigma = cov[X,X] = E[(X - E[X])(X - E[X])^T] = E[XX^T] - E[X]E[X]^T .$$

Approximating the distribution of each class with a normal distribution allows drawing samples of candidate vectors representing an instance of the specific class. Such vectors can replace the values in the embedding's placeholder $e_i^l$ of the specific class $l$ in an image embedding $E_i$. The reconstruction of the modified embedding resembles a possible instance of the dataset that belongs to the specific class. For example, the embedding of an image that does not satisfy the label "*mouth slightly open*" may be modified by inserting a vector $\vec{x}$ sampled from the distribution of the label "*mouth slightly open*" to the specific embedding's placeholder $e^l$ corresponding to the specific label $l$. After making the assignment $e^l = \vec{x}$, passing the modified embedding through the reconstruction model, this creates an image similar to the original which additionally satisfies the specific label. In other words, the face in the image remains very similar but additionally it has the "*mouth slightly open*" property. Respectively, the phenotype of a label can be removed by filling the embedding's placeholder that corresponds to the label with zeros or by replacing the values of the placeholder with a vector sampled from the distribution of the specific label after being scaled down to a small norm value. Vector upscaling can also be applied when adding a specific property to an embedding by replacing a placeholder with a sampled vector. In this way, the effect of adding a specific property is increased and the phenotype change can be more evident.

Figure 7 shows several cases of sampling the embeddings distributions for generating images with specific characteristics or for removing specific characteristics from images. Interestingly, the modified embeddings generate images of faces that are very similar to the reconstructed images when using the original unaltered embeddings. Additionally, the new image has the desired characteristic added to the embedding of the image. This experiment suggests that the *32-d* vector components of $e^l$ encode the various image properties in an effective manner.

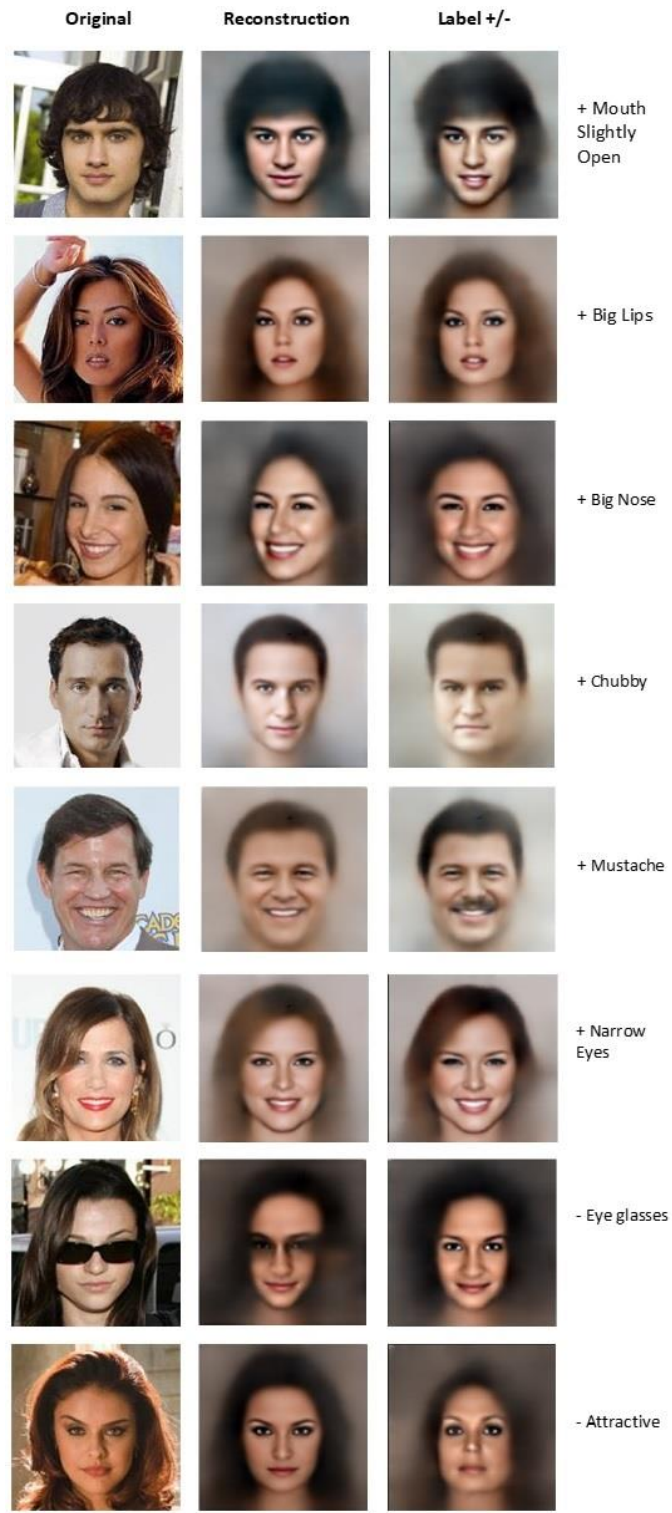

Figure 7. Examples of generating images with a property removed or added. The original image is shown on the left column, the reconstruction of its unmodified embedding in the $2^{nd}$ column and the reconstruction of its modified embedding on the $3^{rd}$ column. The desired property added or removed by modifying the embedding is shown in the right most column. A plus (+) prefix indicates the replacement of the appropriate embedding's placeholder with a vector sampled from the normal distribution of the embeddings that satisfy the specific characteristic (label). A minus (-) prefix indicates the replacement of the appropriate embedding's placeholder with a zero vector. The images belong to the test set.

The degree of an edited characteristic can also be controlled by adjusting the magnitude of the sampled vector. For example, in order to add an emphasized phenotype of a specific label to an image, the magnitude of the sampled vector may be increased by multiplying the vector with a scaling factor $s > 1$. The opposite (reduction of the vector's magnitude) tends to add mild phenotypes. Figure 8 shows that increasing the magnitude of the sampled vectors makes the edited characteristic become more evident.

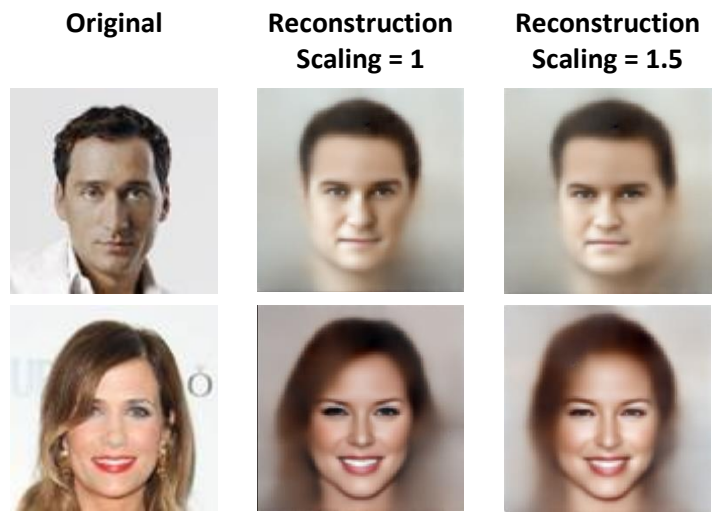

Figure 8. The edited characteristics become more evident when the added vector is scaled by a factor to increase its magnitude. In the first row, a vector is sampled from the *"chuppy"* distribution and used to edit the original image. The right-most image shows the result of editing the embeddings with a vector multiplied with a scaling factor $s = 1.5$. In the second row, the editing vector is sampled from the *"narrow eyes"* distribution.

## 5. Discussion

OLR aims to calculate the representations of each problem label while preserving the evident label correlations. It does not explicitly address attribute-editing nor representing the instances in terms of single-value class probabilities. In a sense, OLR has some similarity with MulGan in the way specific labels get a fixed placeholder in the latent distribution. It is different from AttGan and STGAN in the sense that the latent distribution of our model comprises solely of label representations not containing any other components that encode general image details. Essentially, OLR encodes all image information in the label distributions while preserving the attribute correlations. Moreover, OLR constructs the label distributions without considering an adversarial, a reconstruction or a classification loss. It does this by simply applying a supervision signal sourced from the actual image labels. Avoiding the use of reconstruction loss enables the model to maintain the correlations between the labels and to depart from adapting according to specific image details. Various experiments were performed (see Sections 4.1-4.5), to demonstrate that OLR has a behavior that indicates its understanding of the semantics of the data distribution.

In regard to the experiment 1 (Section 4.1), using a dot product for applying the labeling on the calculated features of the Siamese model is the key operation that differentiates OLR from the conventional way of making the classification with a fully connected classifier. Usually, the classifier consists of a large number of neurons and accepts as input the features detected from the preceding convolutional layers and form complex non-linear relations to satisfy the output labeling. In other words, the fully-connected layers at the end of the conventional classifier combine the calculated features in uncontrolled and arbitrary ways under one criterion: fitting the labels available. On the other hand, the embeddings based on the dot product must satisfy a probabilistic criterion: same labels produce label distributions that are similar in the dot-product sense. The proposed method produces label distributions that have non-zero values only if specific features are detected in images having the same label. In this way, the proposed approach validates its perception on images and its decisions regarding the conformance of each image to a label. This conformance must be "*justified*" in the sense that the compressed content vector of a specific label must have a considerable probability of occurrence in other images having the same label.

Then, Experiment 2 (Section 4.2) suggests that OLR can calculate embeddings that capture the relations between the different problem labels. While these interrelations are seemingly easy for humans to infer, establishing these logical links is not an easy task for machine learning (ML) models. Further, Experiment 3 (Section 4.3) demonstrates that the principal component of the embeddings' set of each label reflects the way the phenotype of the specific characteristic is imprinted on the data.

Moreover, Experiment 4 (Section 4.4) demonstrates that the learned embeddings can be transformed back to the images that produced them. Despite the very small size of the embeddings in comparison to the original data and the discarding of a huge amount of information, the calculated embeddings are still able to maintain enough information to reproduce a descent version of the input. Finally, Experiment 5 (Section 4.5) shows that the proposed method provides label distributions that can be exploited in various ways for semantically-aware image editing: an instance of a characteristic may be sampled from the specific distribution and added to an image while another characteristic may be removed from an image by significantly reducing (or eliminating) the magnitude of the respective distribution. The phenotype of the edited characteristic (characteristic intensity) can be controlled by modifying the magnitude of the sampled instance.

### 5.1 Limitations

While there is nothing restraining the OLR from working with problems that have fewer labels or a single label per image, its full potential unravels when dealing with problems having many labels per image. However, despite OLR uses labels in an indirect way, it is still limited due to its reliance on supervisory information. Another limitation rises from the fact that, during the experiments, the image embeddings were not allowed to be adapted and were used as input data rather than intermediate/learnable features. As such, they did not adapt depending on the task and therefore they were not specialized in tackling the specific problems. On one hand, using unspecialized representations for a variety of tasks stresses their quality, but on the other hand, it produces worse results which makes the task-specific assessment of the model more difficult. For example, a comparison between the results of the attribute-editing experiment (Section 4.5) and the results of analogous models is highly unfair because our experimental model uses embeddings that have been learned without considering the task under study. Future work aspires to address this limitation.

**5.2 Implications and future work**

Finding a latent space at which we are fully aware of what each variable controls is a huge step forward in the research direction of semantically-aware deep learning and computer vision in general. OLR applies latent space factors' disentanglement, which is derived from its architecture and training procedure. Every latent representation has a non-zero magnitude only if its respective label is evident in an image. Occlusion-based supervision drives the model into building representations that reflect its degree of belief that an image complies to a label. More importantly, these representations encapsulate image semantics as suggested by the conditional reconstruction experiment (Experiment 5). Converting labels to meaningful vectors is especially useful in many aspects. Both main ML regimes (supervised and unsupervised learning) can benefit from exploiting the information distilled in label representations. As shown in the experiments conducted, OLR builds label embeddings with appealing properties that may be harvested by ML methods.

As future work, we plan to add an adversarial loss [22] to the training of the decoder to improve the quality of the decoded images. Specifically, a discriminator will be added at the output of the decoder and trained with real and generated images. Optimization of a loss based on both MSE and real/fake adversary has been used before. FSRGAN [24] uses this technique and achieves the current perceptual state-of-the-art in face super resolution task for x8 upscaling. Another interesting direction would be to modify the model for directly outputting distributions instead of plain representations. This would resemble variational autoencoders [23] which use a (normal) distribution for their latent space. Converting the image labels to distributions would be helpful for the attribute editing application in terms of sampling an attribute point and controlling the degree of the attribute phenotype in the generated image (high probability samples of an attribute should produce an emphasized attribute in the output image).

**6. Conclusions**

We have presented a simple method for calculating effective representations of labels that capture the relations among them, using a Siamese network and a dataset of human faces. Several experiments were conducted revealing the potential of the methodology adopted and its ability to provide meaningful label embeddings. The results of the experiments suggest that the small size of the calculated embeddings does not prevent them from maintaining sufficient information regarding the semantics of the data. Moreover, the experiments performed indicate the big potential of methods that transform labels to information-rich vectors.

**Acknowledgments**

The authors have received funding from the European Union's Horizon 2020 Research and Innovation Programme under grant agreement No 739578 complemented by the Government of the Republic of Cyprus through the Directorate General for European Programmes, Coordination and Development.